\documentclass{article}
\usepackage{graphicx} 

\title{Extraction of Research Objectives, Machine Learning Model Names, and
Dataset Names from Academic Papers and Analysis of Their Interrelationships
Using LLM and Network Analysis}

\author{S. Nishio, H. Nonaka, N. Tsuchiya, A. Migita, Y. Banno\thanks{Aichi Institute of Technology}\and  T. Hayashi\thanks{University of Tokyo}\and  H. Sakaji\thanks{Hokkaido University}\and  T. Sakumoto\thanks{Nagaoka University of Technology}\and  K. Watabe\thanks{Saitama University}}

\begin{document}
\maketitle
\begin{abstract}
   Machine learning is widely utilized across various industries. Identifying the appropriate machine learning models and datasets for specific tasks is crucial for the effective industrial application of machine learning. However, this requires expertise in both machine learning and the relevant domain, leading to a high learning cost. Therefore, research focused on extracting combinations of tasks, machine learning models, and datasets from academic papers is critically important, as it can facilitate the automatic recommendation of suitable methods. Conventional information extraction methods from academic papers have been limited to identifying machine learning models and other entities as named entities. To address this issue, this study proposes a methodology extracting tasks, machine learning methods, and dataset names from scientific papers and analyzing the relationships between these information by using LLM, embedding model, and network clustering. The proposed method's expression extraction performance, when using Llama3, achieves an F-score exceeding 0.8 across various categories, confirming its practical utility. Benchmarking results on financial domain papers have demonstrated the effectiveness of this method, providing insights into the use of the latest datasets, including those related to ESG (Environmental, Social, and Governance) data.
\end{abstract}

\section{Introduction}
The use of machine learning for analysis has rapidly spread in recent years and is now employed in various fields such as services and finance \cite{1}\cite{2}
\cite{3}\cite{4}. In this context, selecting the appropriate data and machine learning methods to solve specific problems requires not only knowledge of machine learning but also domain knowledge of the field being analyzed. Therefore, establishing methods to support decision-making has become urgent. There is an increasing amount of research focused on extracting technical terms, such as machine learning methods and dataset names, from academic papers and patent documents to aid in decision-making. For example, an early study before the widespread adoption of deep learning by \cite{5} focused on extracting the technologies used in literature. However, studies conducted before the advent of deep learning faced performance issues and had practical limitations.
Recently, models leveraging deep learning have emerged to improve performance. Studies such as \cite{6} and \cite{7} have proposed methods to extract machine learning methods and dataset names from academic papers using language models. However, to “select the appropriate data and machine learning methods for problem-solving,” it is necessary to comprehensively analyze the relationships between research objectives, datasets, and machine learning methods, rather than merely extracting them. Furthermore, synonymous terms, such as “SVM” and “Support Vector Machine,” need to be recognized as the same expression to avoid separate analyses, which could lead to inconvenient statistical trends. Therefore, methods for semantic aggregation must also be employed.
In this study, we propose a method that utilizes the large language model Llama2 to extract research objectives, machine learning methods, and dataset names from individual papers. The extracted expressions are then aggregated based on synonym relationships using the embedding model E5. Additionally, we analyze the relationships between objectives, machine learning methods, and datasets using network clustering based on co-occurrence graphs within the papers. This approach provides valuable information for making decisions about the appropriate data and machine learning methods to use for problem-solving. Moreover, the results of this method could be applied in the future for the automatic recommendation of datasets and machine learning models. We demonstrate the practical applicability of our method by evaluating and analyzing its performance in the field of quantitative finance, where numerous papers are freely available on arXiv. The evaluation and analysis focus on the frequency of extracted expressions and the relationships between datasets, as well as significant connections between objectives and datasets.

\begin{figure}
    \centering
    \includegraphics[width=0.5\linewidth]{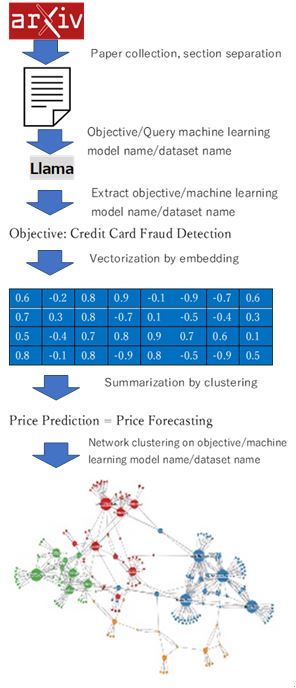}
    \caption{Research Overview Diagram}
    \label{fig:in-label}
\end{figure}

\section{Methodology}
\subsection{Overview of the Proposed Method}
This section provides an overview of the proposed method. An outline diagram is shown in Figure \ref{fig:in-label}. First, the large language model Llama is used to simultaneously extract research objectives, machine learning methods, and dataset names. Then, to address variations in expressions due to synonyms and supplementary explanations using parentheses, clustering is performed using embeddings with E5. Subsequently, the clusters are treated as nodes, and a graph network is constructed with edges based on co-occurrence relationships within the papers. Network clustering of this graph is used to analyze the interrelationships between objectives, datasets, and machine learning model names. The details are described below.

\subsection{Extraction of Research Objectives, Machine Learning Model Names, and Dataset Names from Papers Using Llama}
In this study, we employ Llama2 and Llama3, opensource models known for their high performance in various natural language processing tasks \cite{8}, for expression extraction. Specifically, we utilize Llama2- 13B and Llama3-8B and provide prompts to extract research objectives, machine learning model names, and dataset names from the body of academic papers. An example of the prompts used for dataset name extraction is shown below. Llama2 and Llama3 are state-of-the-art open-source language models that have demonstrated superior performance across a wide range of natural language processing (NLP) tasks. These models are part of the Llama family, which is known for its scalability and effectiveness in handling complex language understanding and generation tasks. 

Llama2: Llama2, specifically the Llama2-13B variant used in this study, is a highly advanced model consisting of 13 billion parameters. This extensive parameterization allows the model to capture intricate linguistic patterns and nuances, making it exceptionally proficient in tasks such as text summarization, question answering, and named entity recognition. Llama2's architecture is built on transformer-based models, which leverage self-attention mechanisms to process and generate human-like text with high accuracy and coherence. 

Llama3: Llama3 represents the next generation in the Llama series, with the Llama3-8B variant employed in this research. Despite having fewer parameters (8 billion) compared to Llama2-13B, Llama3 introduces several architectural enhancements and optimizations that improve its computational efficiency and performance. These improvements include advanced training techniques, better handling of long-range dependencies, and refined language modeling capabilities. Llama3's design focuses on achieving a balance between model size and performance, making it an effective tool for various NLP applications while maintaining lower computational costs.

Both Llama2 and Llama3 are designed to be adaptable and can be fine-tuned for specific tasks, which enhances their utility in domain-specific applications. Their opensource nature facilitates widespread adoption and customization, enabling researchers and developers to build on their robust foundational capabilities. In this study, the integration of Llama2-13B and Llama3-8B models allows for precise extraction of research objectives, machine learning model names, and dataset names from academic papers, demonstrating their practical application in automating and enhancing research workflows.

\begin{figure}
    \centering
    \includegraphics[width=0.5\linewidth]{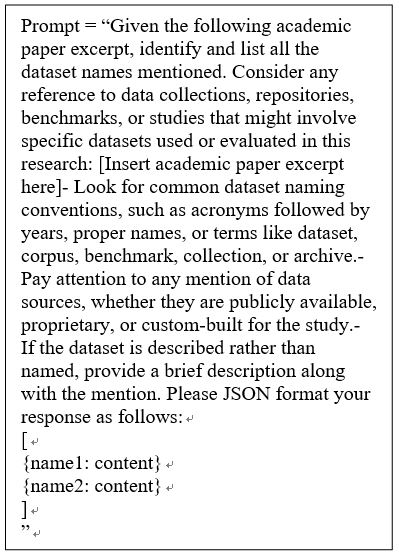}
    \caption{Prompt used for data set name extraction}
\end{figure}

To avoid redundancy and reduce computational load, we limit the extraction to specific sections of the papers. For instance, when extracting research objectives, we focus solely on the Introduction section, thereby improving processing speed. By targeting specific sections, we enhance both the accuracy and efficiency of the extraction process. For the extraction of research objectives, we use prompts designed to identify responses to questions such as "What is the purpose of this study?" primarily within the Introduction section. For extracting the names of machine learning models, we set prompts to identify mentions of models likely found in sections such as Methods or Results, with questions like "Which models are used in this study?" For dataset name extraction, we target sections such as Data or Experiments, using prompts like "Which datasets are used?" This targeted approach allows us to avoid unnecessary data processing and efficiently extract the required information, thereby saving computational resources and improving the quality of the extraction results.

\subsection{Clustering Using Embedding Models}
The expressions extracted above are consolidated into synonymous expressions. In this study, we vectorize words and sentences based on their meanings and cluster similar ones to aggregate synonymous expressions. Additionally, supplementary explanations using parentheses can become noise, but clustering helps to consolidate these noises as well. We use E5, an embedding model that has demonstrated excellent performance in various tasks \cite{9}. For clustering the distributed representations, we adopt Ward's method, which is widely used in hierarchical clustering.

\begin{figure}
    \centering
    \includegraphics[width=0.6\linewidth]{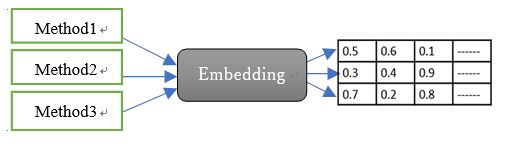}
    \caption{Image of embedding}
\end{figure}

\begin{figure}
    \centering
    \includegraphics[width=0.5\linewidth]{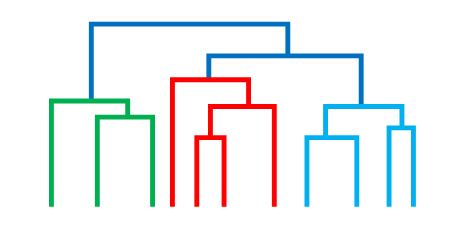}
    \caption{Image of clustering}
\end{figure}

\subsection{Creation of Co-occurrence Graph and Network Clustering} After consolidating synonyms, we construct a graph network by connecting co-occurring nodes in the papers as edges, with each cluster serving as a node. This clarifies the relationships between research objectives, machine learning methods, and datasets, enabling visualization and analysis of which combinations of machine learning methods and datasets are used for specific objectives within each domain. This approach aids in decision-making for the application of machine learning within a domain and could potentially lead to the recommendation of appropriate datasets and machine learning models based on the identified tasks. Additionally, we performed grouping using the network clustering method by Girvan and Newman algorithm which is a method for detecting communities by iteratively removing edges with the highest betweenness centrality, thus progressively partitioning the network. This algorithm is one of the fundamental approaches in network clustering and has been widely used in numerous studies.

\begin{figure}
    \centering
    \includegraphics[width=0.4\linewidth]{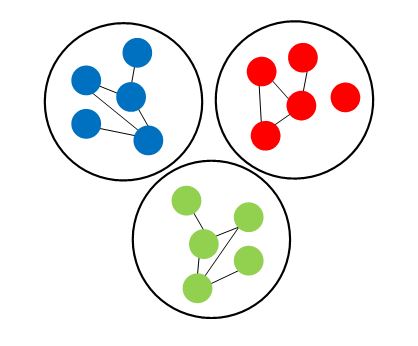}
    \caption{Image of Network clustering}
\end{figure}

\section{Result}
In this study, we collected and analyzed arXiv papers related to econometrics. We used the Python libraries ‘request’ and ‘arxivAPI’ for paper collection. The search criteria for collection involved specifying "quantitative finance" and targeting papers related to the use of machine learning models and datasets. We conducted an "and" search using the terms "Machine Learning" and "Dataset," resulting in the collection of 181 papers. We utilized Llama and E5 via the Python library ‘langchain’ for processing. An example of actual results extracted using Llama is shown below. The yellow markers in the paper are the extracted parts.

An evaluation experiment using Llama was conducted for information extraction. Each item extracted from a randomly selected set of 10 papers was evaluated. The evaluation was performed by two individuals: one with expertise in natural language processing and the student. For all items, Llama3 outperformed Llama2.The evaluation results of information extraction by Llama2 and Llama3 are shown in the table \ref{tab:my_label} below.

Additionally, we employed the Python libraries `sqlite3` and `faiss` for database management and clustering, respectively. Clustering was performed using the Python library `Scipy`, and co-occurrence graph analysis was conducted using the Python library `Networkx`. These libraries were implemented on Google Pro+. Below is the co-occurrence graph of 3 elements and the co-occurrence graph for objective-dataset.

\begin{figure}
    \centering
    \includegraphics[width=0.6\linewidth]{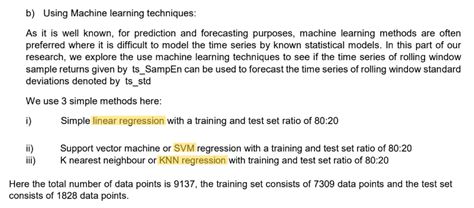}
    \caption{Machine learning models actually extracted from the paper}
\end{figure}

\begin{table}
    \centering
    \begin{tabular}{|c|c|c|c|} \hline 
         &  Dataset&  Method& Objective\\ \hline 
         Llama2&  0.642&  0.854& 0.871\\ \hline 
         Llama3&  0.870&  0.935& 0,955\\ \hline
    \end{tabular}
    \caption{Evaluation experiment results (F1 score)}
    \label{tab:my_label}
\end{table}

\begin{figure}
    \centering
    \includegraphics[width=0.4\linewidth]{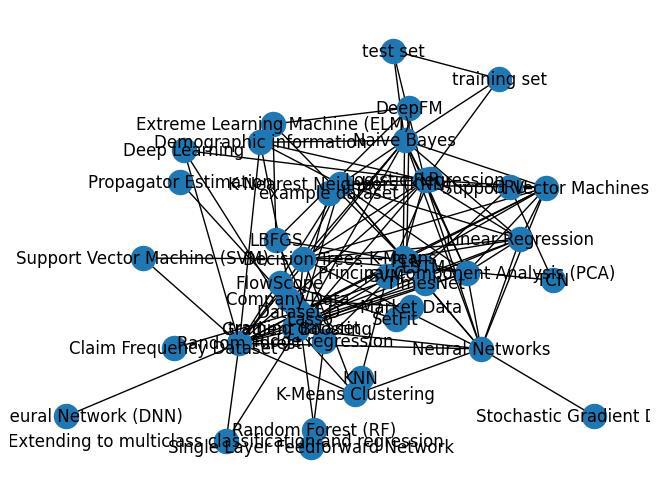}
    \caption{Co-occurrence Graph of 3 Elements}
    \label{fig:re-label}
\end{figure}

\begin{figure}
    \centering
    \includegraphics[width=0.4\linewidth]{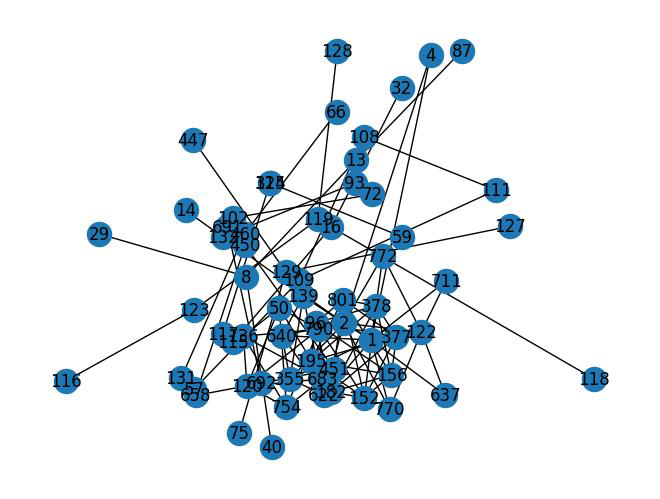}
    \caption{Co-occurrence Graph of Objectives and Datasets}
\end{figure}

\section{Discussion}
Upon extracting dataset names for the field of quantitative finance, we found that stock price data, such as SP500, emerged as the most frequent, indicating a vibrant research landscape utilizing stock price data. The results of the evaluation experiment show that the dataset item scored lower compared to other items. This is likely because the extracted results included datasets uniquely created by the authors of the papers, as well as training and test datasets used for the learning and evaluation of models. However, the results from the cooccurrence graph in Figure \ref{fig:re-label} and the network clustering revealed a group of studies utilizing text data from platforms like Stocktwits, focusing on research tasks such as utilizing datasets aligned with social themes like ESG or predicting economic indicators like stock prices. While these studies had lower frequency, they signify emerging research trends, suggesting the utility of this study in discovering new avenues of research.

\section{Conclusion}
In this study, we proposed a method for extracting research objectives, machine learning models, and dataset names simultaneously, followed by analyzing their interrelationships using co-occurrence graphs and network clustering. As a result, we demonstrated the capability of visualizing research trends in the field of quantitative finance. The proposed method's expression extraction performance, when using Llama3, achieves an F-score exceeding 0.8 across various categories, confirming its practical utility. Benchmarking results on financial domain papers have demonstrated the effectiveness of this method, providing insights into the use of the latest datasets, including those related to ESG (Environmental, Social, and Governance) data.

\bibliographystyle{alpha}
\bibliography{ref}

\newcommand{\etalchar}[1]{$^{#1}$}
\begin{thebibliography}{ACMENH21}

\bibitem[ACMENH21]{1}
E.~C. Alemán~Carreón, H.~A. Mendoza~Espa˜na, H.~Nonaka, and T.~Hiraoka.
\newblock Differences in chinese and western tourists faced with japanese hospitality: a natural language processing approach.
\newblock {\em Information Technology Tourism}, 23:381--438, 2021.

\bibitem[HMPM21]{6}
J.~Heddes, P.~Meerdink, M.~Pieters, and M.~Marx.
\newblock The automatic detection of dataset names in scientific articles.
\newblock {\em Data}, 6(8):84, 2021.

\bibitem[KWS{\etalchar{+}}21]{4}
H.~Kamimura, J.~Watanabe, T.~Sugano, J.~Kohisa, H.~Abe, K.~Kamimura, M.~Takamura, S.~Okoshi, Y.~Tanabe, R.~Takagi, H.~Nonaka, and S.~Terai.
\newblock Relationship between detection of hepatitis b virus in saliva and periodontal disease in hepatitis b virus carriers in japan.
\newblock {\em Journal of infection and chemotherapy}, 27(3):492--496, 2021.

\bibitem[Nea18]{2}
K.~Nakai and et~al.
\newblock Community detection and growth potential prediction using the stochastic block model and the long short-term memory from patent citation networks.
\newblock {\em In 2018 IEEE International Conference on Industrial Engineering and Engineering Management (IEEM)}, pages 1884--1888, 2018.

\bibitem[NKS{\etalchar{+}}12]{5}
H.~Nonaka, H.~Kobayashi, A.and~Sakaji, Y.~Suzuki, H.~Sakao, and S.~Masuyama.
\newblock Extraction of effect and technology terms from a patent document (theory and methodology).
\newblock {\em Journal of Japan Industrial Management Association}, 63(2):105--111, 2012.

\bibitem[Tea23]{8}
H.~Touvron and et~al.
\newblock Llama 2: Open foundation and fine-tuned chat models.
\newblock {\em arXiv preprint}, page arXiv:2307.09288, 2023.

\bibitem[Wea22]{9}
L.~Wang and et~al.
\newblock Text embeddings by weaklysupervised contrastive pre-training.
\newblock {\em arXiv preprint}, page arXiv:2212.03533, 2022.

\bibitem[YN21]{3}
H.~Yamashiro and H.~Nonaka.
\newblock Estimation of processing time using machine learning and real factory data for optimization of parallel machine scheduling problem.
\newblock {\em Operations Research Perspectives}, 8:100196, 2021.

\bibitem[YYZ{\etalchar{+}}23]{7}
R.~Yao, Y.~Ye, J.~Zhang, S.~Li, and O.~Wu.
\newblock Exploring developments of the ai field from the perspective of methods, datasets, and metrics.
\newblock {\em Information Processing Management}, 60(2):103157, 2023.

\end{thebibliography}

\end{document}